\documentclass[information,article,accept,pdftex,moreauthors]{Definitions/mdpi} 

\usepackage[labelformat=simple]{subcaption}

\DeclareCaptionLabelFormat{subcaptionlabel}{\normalfont\hspace{50pt}(\textbf{#2}\normalfont)}
\usepackage{amsthm}  
\usepackage{listings}
\usepackage{amssymb}
\usepackage{latexsym}
\usepackage{epsfig}
\usepackage{float}
\usepackage{gensymb}
\usepackage{xspace} 
\firstpage{1} 
\pubvolume{1}
\issuenum{1}
\articlenumber{0}
\pubyear{2023}
\copyrightyear{2023}
\externaleditor{Academic Editor: Francesco Camastra %MDPI: please add the names.
}
\datereceived{12 April 2023 } 
\daterevised{13 June 2023 } % Comment out if no revised date
\dateaccepted{20 June 2023 } 
\datepublished{ } 
\hreflink{https://doi.org/} % If needed use \linebreak
\Title{Reef-Insight: A Framework for Reef Habitat Mapping  with Clustering Methods Using Remote Sensing%Attention: title altered. Check intended meaning retained.
}

\TitleCitation{Reef-Insight: A Framework for Reef Habitat Mapping  with Clustering Methods Using Remote Sensing}

\Author{ {Saharsh Barve} %MDPI: Please carefully check the accuracy of names and affiliations.
 $^{1}$,  {Jody M. Webster} %MDPI: The name is different from system, please confirm which one is correct. CORRECT
 $^{2}$ and Rohitash Chandra $^{3,}$*}

\AuthorNames{Saharsh Barve, Jody Webster and Rohitash Chandra}

\AuthorCitation{ {Barve, S.;} %MDPI: Please carefully check the accuracy of names.
 Webster, J.M.; Chandra, R.}
\address{%
$^{1}$ \quad Department of Computer Science and Engineering, Manipal Institute of Technology,  {Manipal,} %MDPI: please add post code in all affs.
 India; saharshbarve3@gmail.com \\
$^{2}$ \quad Geocoastal Research Group, School of Geosciences, University of Sydney,  {Sydney,} 
 Australia; jody.webster@sydney.edu.au\\

$^{3}$ \quad Transitional Artificial Intelligence Research Group, School of Mathematics and Statistics, UNSW Sydney,   {Sydney,}
 Australia}

\corres{Correspondence: rohitash.chandra@unsw.edu.au}
 
\abstract{Environmental damage has been of much concern, particularly in coastal areas and the oceans, given climate change and the drastic effects of pollution and extreme climate events.  Our present-day analytical capabilities, along with advancements in information acquisition techniques such as  remote sensing, can be utilised for the management and study of coral reef ecosystems. In this paper, we present Reef-Insight,  an unsupervised machine learning framework that features  advanced clustering methods and remote sensing  for reef habitat mapping. Our framework compares different clustering methods  for reef habitat mapping using remote sensing data. We evaluate four major clustering approaches based on qualitative and visual assessments which include k-means,  hierarchical clustering, Gaussian mixture model, and density-based clustering. We utilise remote sensing data featuring  the One Tree Island  reef in Australia's Southern Great Barrier Reef.  Our results indicate that clustering methods using remote sensing data can well identify  benthic and geomorphic clusters  in reefs  when compared with other studies. Our results indicate that Reef-Insight can  generate  detailed reef habitat maps outlining  distinct reef habitats and has the potential to enable  further insights for reef restoration projects.  }

\keyword{clustering; reef habitat mapping;  Great Barrier Reef; machine learning} 
 
\begin{document}

%%%%%%%%%%%%%%%%%%%%%%%%%%%%%%%%%%%%%%%% 

\section{Introduction}
   
 Remote sensing provides the methodology that enables aerial data to be retrieved using advanced satellites and aerial vehicles~\citep{navalgund2007remote,pajares2015overview}. In recent decades, remote sensing has been prominent in a number of applications, which include  tropical forest environmental  monitoring~\citep{foody2003remote}, environmental monitoring ~\citep{li2020review}, mining environmental monitoring~\citep{padmanaban2017remote}, coral reef monitoring~\citep{hedley2016remote}, agriculture~\citep{seelan2003remote}, surface moisture and soil monitoring~\citep{wang2009satellite},  and space research~\citep{elachi1988spaceborne}. Remote sensing data  with machine learning methods have been increasingly used~\citep{lary2016machine,maxwell2018implementation,ali2015review} in diverse applications, such as mineral exploration~\citep{shirmard2022review}, environmental monitoring, and agriculture~\citep{holloway2018statistical}.

Coral reef mapping~\citep{mumby1997coral,hamylton2017mapping} provides valuable information about reef characteristics, such as the structure of the reef,  geomorphic and benthic zones, and coral distribution, which can help in reef restoration projects~\citep{bayraktarov2019motivations,jaap2000coral}. Some of the related studies that used remote sensing are discussed as follows:  Kennedy et al.~\citep{kennedy2021reefcover} proposed a  coral classification system that combines satellite data and local knowledge for identifying different geomorphic regions in a coral reef. Among various analytical techniques used on remote sensing data,  Phinn et al.~\citep{phinn2012object} assessed the quality of   benthic and geomorphic   community maps of coral reefs produced with multi-scale  image analysis. Other map-processing approaches commonly used are supervised classification~\citep{white2021cr-map-rs} and manual delineation of classes using images as a backdrop. Phinn  et al.~\citep{roeflsema2009eight} evaluated eight commonly used benthic cover mapping techniques based on the two processing approaches stated above. Eight techniques were assessed on the basis of cost, accuracy, time, and relevance, where the preferred mapping approach was supervised learning using  the classification  of satellite data using basic field knowledge. Nguyen et al.~\citep{nguyen2021map}  provided a review of coral reef mapping with multispectral satellite   image correction, and pre-processing techniques and classification algorithms.  
Machine learning enables valuable information to be extracted from remotely sensed data with   clustering~\citep{zhang2016spectral}, dimensionality reduction~\citep{shirmard2020integration}, and classification methods~\citep{maxwell2018implementation}. Machine learning methods are slowly becoming prominent for  climate change problems~\citep{rolnick2022tackling}. These methods  can also be used to understand and reconstruct data for climate and vegetation   millions of years back in time~\citep{chandra2021precipitation}.  Clustering  ~\citep{bandyopadhyay2007multiobjective} is an unsupervised machine learning method that is useful for remote sensing   when labelled  data are unavailable. The clusters produced using clustering  techniques can be further improved  using specific spatial information from the data   or by applying basic domain knowledge  ~\citep{v2013kmeans,Chakraborty20}.  Clustering techniques   can be used for image segmentation tasks~\citep{naik2014review}, where pixels are grouped into distinct regions (clusters) on the basis of given similarity measures. There are several clustering techniques that serve this purpose, such as the k-means  clustering algorithm~\citep{Macqueen67somemethods}, agglomerative hierarchical clustering~\citep{Johnson_hac}, density-based spatial clustering (DBSCAN)~\citep{ester1996densityx}, and Gaussian mixture model (GMM)~\citep{Biernacki_GMM}. %Each of these has specific properties which make them more useful than others  for remote sensing data for different applications, and their strengths and weakness in the area of coral reef mapping is yet to be evaluated. 

{Our choice of clustering methods is based on their properties for the segmentation of image-based data. Our literature review indicates that the selected methods have strengths and limitations in various applications; however, our study evaluates them to segment reef community mapping based on satellite-based image data.}

 In the case of hyperspectral and multispectral data that feature multiple bands and thus feature a large number of data, dimensional  reduction methods, such as principal component analysis (PCA)~\citep{zabalza2014novel}, can   reduce the number of  bands  in order to make the data applicable  for clustering methods. They have been used in remote sensing applications, such as mineral exploration with satellite data~\citep{shirmard2020integration}.
Existing work conducted in creating reef maps using supervised learning techniques can be utilised for qualitative comparisons with clustering results and for labelling the regions. The success of clustering techniques in mapping regions and in the field of geosciences inspires the use of the technique for the benthic~\citep{hochberg2000spectral} and geomorphic mapping~\citep{locker2010geomorphology} of coral reefs. Although remote sensing data (multispectral and hyperspectral) have been   used for coral reef mapping~\citep{holden1999hyperspectral,lesser2007bathymetry,philipson2003can,goodman2013coral,zhang2015applying,hedley2016remote}, not much has been conducted using   clustering techniques, particularly using  open source software frameworks. There are proprietary remote sensing software suits~\citep{su12135237,selgrath2016mapcoral} that have inbuilt features for reef mapping, but these are not easily available.  This is a problem not just for reproducible research, but for the application of such methods in developing countries, where such software suites are not economically viable for research purposes, and this slows down research and development in the area of reef monitoring, which is a major focus of climate change-related research. {The main problem in reef community mapping is to automatically detect different communities of coral systems, which is challenging with limited, noisy and sparse datasets.  }

%aim - goals 
In this paper, we present an unsupervised machine learning framework using novel clustering methods for the detection and mapping of coral reef habitats with remote sensing.
We present Reef-Insight, a framework for reef community mapping using remote sensing with which we compared  four major clustering approaches in order to determine which method is the most suitable based on qualitative and visual assessments. The clustering methods include k-means, GMM, agglomerative clustering, and DBSCAN. We utilise a remote sensing dataset from One Tree Island   in the Southern Great Barrier Reef in Australia.   Our framework provides the detection and  generation of detailed maps that highlight distinct reef habitats that can guide  scientists and policymakers in  reef restoration projects. Our key innovation is in the development of a framework that can take different forms of data (benthic and geomorphic maps) and evaluate prominent clustering methods for reef community mapping.

 % Structure of the paper
 The rest of the paper is organised as follows: In Section~\ref{sec2}, we present the proposed methodology, followed by experiments and results in Section~\ref{sec3}. Section~\ref{sec4} provides a discussion, and Section~\ref{sec5} provides the conclusion with directions for future work. 

\section{Methodology}\label{sec2}

\subsection{Study Area} 
\begin{figure}[H]

  \hspace{-5pt}  \includegraphics[width=13cm]{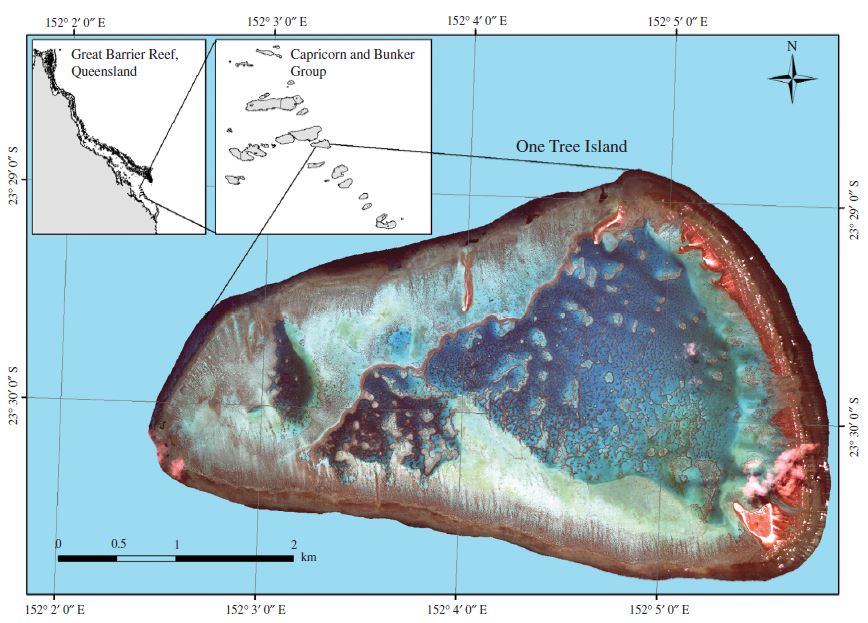}

    \caption{ {One} %MDPI: please replace with sharper image. This is all we got 
 Tree Island reef  {(located at 23 \degree 30$'$ 30$''$ south, 152 \degree 05$'$ 30$''$ east) in the Capricorn–Bunker group} in relation to the Queensland coastline (taken from~\citep{Hamylton2013reef-compare}).}
    \label{fig:otr_location}
\end{figure}

One Tree Island (located at 23 \degree 30$'$ 30$''$ south, 152 \degree 05$'$ 30$''$ east) is a coral reef in the southern Great Barrier Reef. It is a part of the Capricorn–Bunker group about 90~km east of Port Gladstone in Queensland, Australia. The University of Sydney maintains the research station on the island, and as such, the One Tree Island reef has been the subject of detailed biological and geological investigation over the past four decades (see~\citep{doo2017spatial,barrett2017reef,sanborn2020new}), including  studies using remote sensing~\citep{shannon2012evolution}. Hence,  the reef habitats and geomorphic zones characterising the One Tree Island reef have been well studied. Our study area (Figure~\ref{fig:otr_location}) is located at the eastern end of a coral reef that spans about  {5.5 km} 
 in length and 3.5 km in width.  %MDPI: we removed kilometres, please confirm . OK
\vspace{-5pt}

\subsection{Data}

\begin{figure}[H]

    \includegraphics[width=10cm]{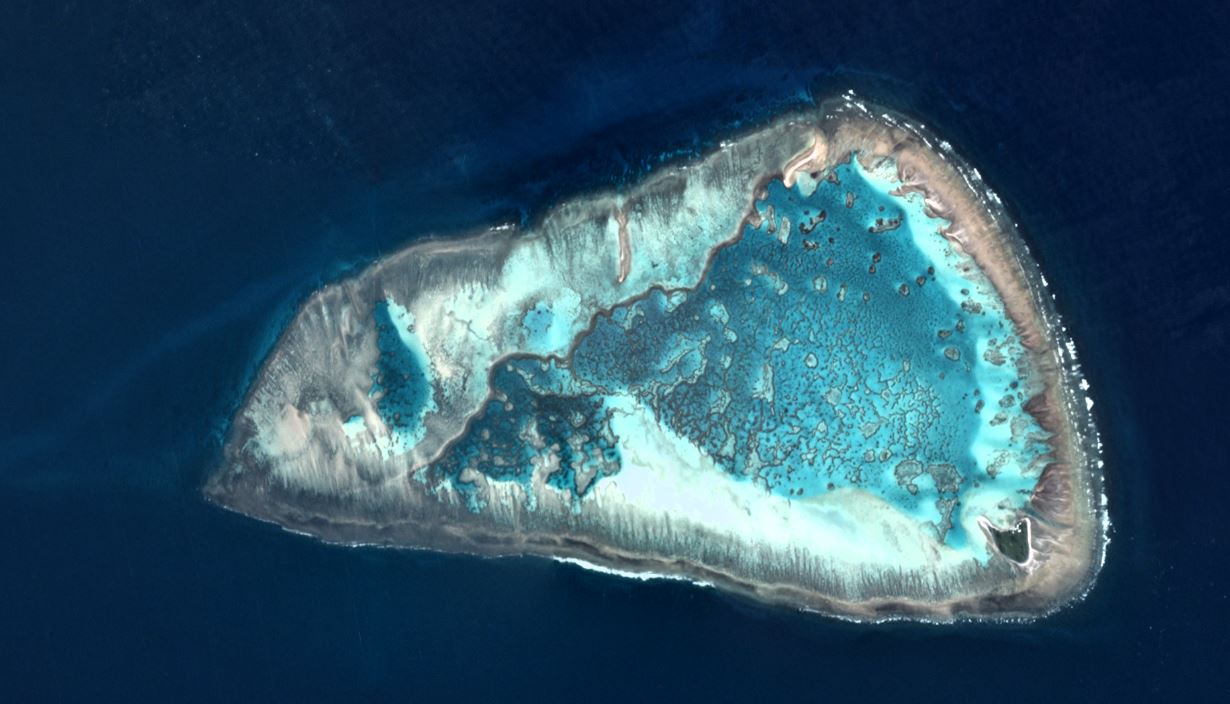}
    \caption{Mosaic of One Tree Island reef taken from the  Allen Coral Atlas~\citep{CoralAtlas-paper}.}
    \label{fig:aca_mosaic}
\end{figure}

We utilise the PlanetScope satellite imagery which is available on the Allen Coral Atlas website. The PlanetScope (Dove)  {(}\url{https://earth.esa.int/eogateway/missions/planetscope},  {accessed on 1st June 2023} {)} %MDPI: 1. footnote is not allowed, so we moved here, please confirm; 2. Please add the access date (format: Date Month Year), e.g., accessed on 1 January 2020. same meanings as below    OK
 image-based  data feature 3 spectral bands (red, green, and blue) at 8-bit radiometric resolution. The raw images are processed for atmosphere radiance, sensor and radiometric calibration, flatfield correction, debayering, orthorectification, and surface reflectance. Furthermore,  mosaic-based processing is done to utilise the “best scene on top” technique   to create the final mosaic. 
The   mosaic process  is taken from the implementation in the Allen Coral Atlas  ~\citep{CoralAtlas-paper}  (Figure~\ref{fig:aca_mosaic}). 
We create bathymetric image data using 10 m resolution  with the Sentinel-2 surface reflectance dataset using  Google Earth Engine (GEE)   {(}\url{https://developers.google.com/earth-engine/datasets/catalog/sentinel-2},  {accessed on 1st June 2023} {)}. Finally, we create a  single mosaic (16-bit integer)  by aggregating the median value of the input data over a period of 12 months. We utilise this bathymetric information for creating geomorphic maps.

  \subsection{Benthic and Geomorphic Regions in Reef}
 
 The benthic~\citep{hochberg2000spectral} mapping of coral reefs refers to the use of  aerial imagery, underwater photos, acoustic surveys, and data  from sediment samples. The benthic zone refers to an ecological region with a low level of water, such as an ocean and coral reef community, that includes the sediment surface.  Geomorphology refers to the evolution of topographic and bathymetric features through physical, chemical, and biological processes on the Earth's surface \citep{summerfield2014global}. Geomorphic coral reef mapping~\citep{locker2010geomorphology} refers to topographic and bathymetric features of reef habitats~\citep{zawada2009multiscale}. 
 
\subsection{Clustering Techniques }

\subsubsection{K-Means Clustering }
%K-means
K-means clustering is an algorithm that divides data into a set of   clusters ($k$) based upon a distance metric \citep{kaufman1990cluster,Macqueen67somemethods}. Given a d-dimensional vector for a dataset of samples ($x_1, x_2, \ldots, x_N$) of size $N$,  the algorithm partitions (groups)  the  data   into $k$ ($\leq$$N$) sets \mbox{$C = C_1 ,C_2, \ldots, C_k$}. The aim of the algorithm is    to minimize the error given by  the within-cluster sum of squares  (WCSS), which is given as the sum of squared  Euclidean distances between the data samples   and the corresponding centroid in the original algorithm ~\citep{hartigan1979algorithm}.
\begin{equation}
    W(C_k) = \sum_{x_i\epsilon C_k}^{1}(x_i - \mu_k)^{2}
\end{equation}
where $x_i$ is a data sample belonging to cluster $C_k$ and $\mu_k$ is the mean  of the samples in  cluster $C_k$. We assign   each data sample    to a given cluster such that the WCSS error  to their assigned cluster centres, $\mu_k$, is minimised. The total  WCSS error is given as follows:
\begin{equation}
  WCSS = \sum_{k=1}^{k}W(C_k) =\sum_{k=1}^{k} \sum_{x_i\epsilon C_k}^{1}(x_i - \mu_k)^{2}
\end{equation}

Although k-means clustering has been prominent in tabular data, it can also be used on image and remote sensing data for segmentation, which is also the focus of this paper. There have been  applications of k-means clustering for remote sensing-based image segmentation, change detection, and land cover classification. Theiler et al.~\citep{theiler_kmeans} proposed a variation of the k-means algorithm to utilise both the spectral and spatial properties of satellite imagery for  image segmentation.  Lv et al.~\citep{lccd_novel}   integrated k-means clustering  with an adaptive majority voting model for land cover change detection.     Celik ~\citep{Celik_ChangeDetect} used dimensional reduction with  PCA and k-means clustering   for the task of change detection. Abbas et al.~\citep{Abbas2016KMeansAI} utilised k-means and ISODATA~\citep{ball1965isodata} (which is an extension of k-means clustering)  for land cover classification using remote sensing data. These applications motivate the use of k-means clustering in our proposed framework.

\subsubsection{GMM}
  
A GMM is based on a probabilistic model that assumes that data are generated from a mixture of   Gaussian distributions with parameters that are adjusted by training. A GMM is useful for clustering, anomaly detection, and density estimation~\citep{Baid2016/12}. It consists of three parameters, which include mean  {($\mu$),} %MDPI: please confirm the change. OK
 which defines the centre of each of the Gaussian distributions%Check intended meaning retained.
 ; covariance  {($\Sigma$)}, which represents the spread; and mixing probability  {($\Pi$)}, defining the weight of the respective Gaussian distribution. The mixing coefficients for each cluster ($k$) are themselves probabilities ($\pi_k$), and must have a sum of 1 as shown below.
\begin{equation}
    \sum_{k=1}^{k}\pi_k =1
\end{equation}

In comparison to GMMs, k-means clustering places a circle (a hypersphere in the case of    higher dimensions) at the centre of each cluster. We can define a radius  with the most distant point in the cluster; however,  GMMs can also handle  oblong and ellipsoidal forms of clusters. The applications of GMMs in remote sensing data processing include image clustering, segmentation, and synthetic data generation. Bei  et al.~\citep{Zhao2016ASG} presented an improvised GMM that takes into account spatial information   to improve  image clustering.  Yin et al.~\citep{Yin_comboGMM} combined the fuzzy region competition method with a GMM for image segmentation. Davari et al.~\citep{Davari_HSRS} utilised a GMM for hyperspectral remote sensing  that featured the challenge of  large dimensions (features) with fewer training data points. Neagoe et al.~\citep{Neagoe_SSclassify} presented a cascade of k-means clustering and a GMM for semi-supervised classification.

\subsubsection{Agglomerative Clustering}

Agglomerative hierarchical  clustering, also known as agglomerative nesting (AGNES), is the most common type of hierarchical clustering used to group data samples in clusters based on their similarity~\citep{johnson1967hierarchical,rohlf1970adaptive,milligan1979ultrametric}. The algorithm begins by treating each data instance as a singleton cluster, and pairs of clusters are    merged until all clusters have been merged into a large  cluster featuring all the data. This produces a dendrogram, which  is a tree-based representation of the data samples.   It produces a flexible and informative cluster tree instead of forcing users to choose a particular number of clusters, such as  determining $k$ in the k-means algorithm. Goncalves et al.~\citep{Goncalves_SOM-HAC} proposed an unsupervised clustering method combining  self‐organising maps (SOMs) with AGNES for the classification of remotely sensed data. Liu et al.~\citep{Liu_HAC-Segment} used hierarchical clustering for the image segmentation of high-resolution remote sensing images.

\subsubsection{DBSCAN}

 DBSCAN~\citep{ester1996densityx}  uses local density estimation to identify clusters of arbitrary shapes, which is not easily possible with traditional methods, such as  k-means clustering. In DBSCAN, the data samples  are seen as core points (density), reachable points, and outliers. The algorithm counts how many samples are located within a small distance   %MDPI: please check this change, if (epsilon) should be removed. OK removed
 from each core point and marks a region  called the  neighbourhood.  The data samples in the neighbourhood of a core sample belong to the same cluster. This neighbourhood may include other core instances; therefore, a long sequence of neighbouring core instances forms a single cluster. Any sample that is not a core sample and does not have one in its neighbourhood is considered an anomaly.   DBSCAN clustering has been prominent in a number of applications with tabular data~\citep{khan2014dbscan} and has been used for remote sensing data. Wang et al. ~\citep{wang2019improved} presented an improved DBSCAN  method for Lidar data,   and the results showed that it could segment different types of  point clouds with higher accuracy in a robust manner. Liang Zhang et al.~\citep{Liang_SuperpixelDBSCAN} utilised DBSCAN clustering in their adaptive superpixel generation algorithm for synthetic-aperture radar (SAR) imagery. Liujun Zhu et al.~\citep{Liujun_soilchange} used  DBSCAN  for vegetation change detection using multi-temporal analysis.

\subsection{Framework}

\begin{figure}[htbp!]

    \includegraphics[width=6.5cm]{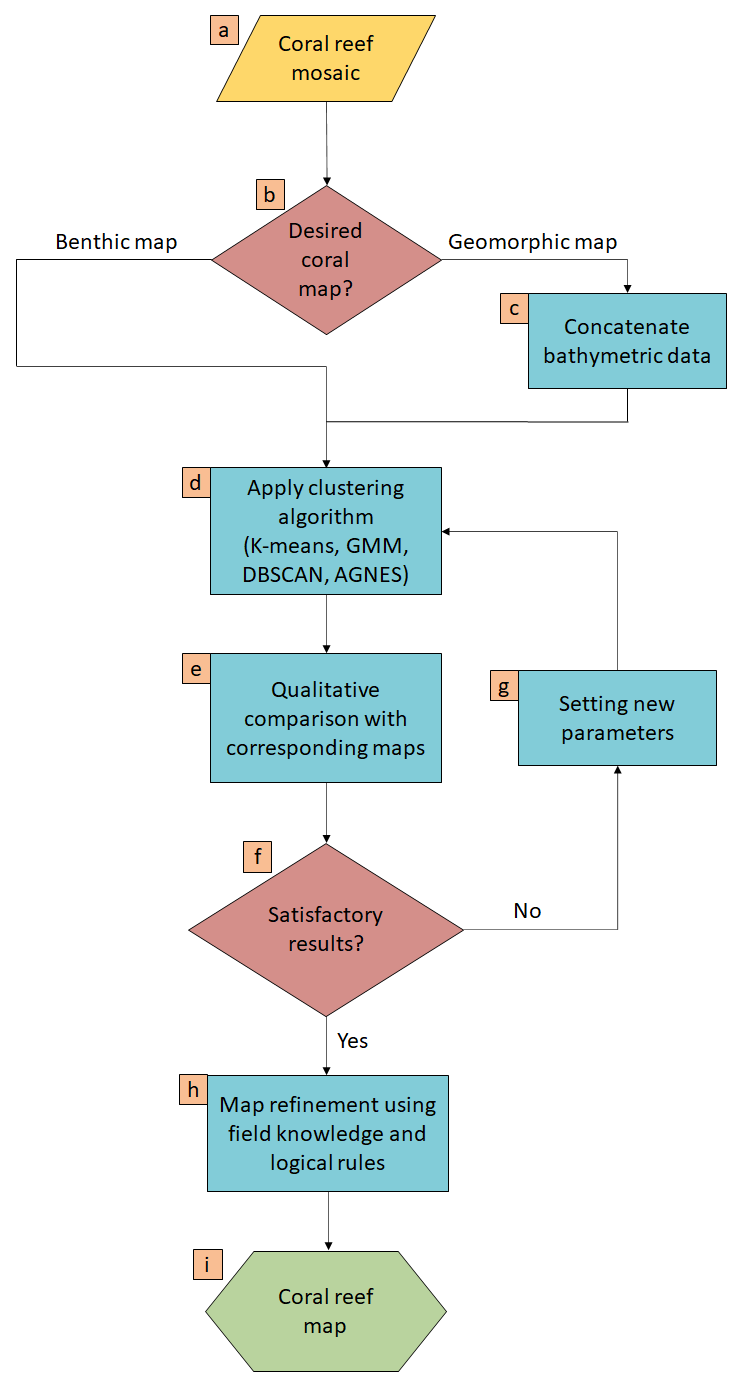}
    \caption{Proposed framework for applying clustering methods  (k-means, GMM, AGNES, and DBSCAN) to remote sensing data for coral reef mapping.}
    \label{fig:flowchart}
\end{figure}

Next, we present the framework that incorporates the different clustering algorithms, i.e., k-means, GMM, agglomerative clustering, and DBSCAN, for the segmentation of two different types of satellite imagery to create coral reef maps (Figure~\ref{fig:flowchart}). The initial step is to acquire remote sensing-based imagery of the coral region of interest (\mbox{Figure~\ref{fig:flowchart}---Step a}). The coral reef mosaic data  taken from the Allen Coral Atlas
 \citep{CoralAtlas-paper} utilise  sensor and radiometric calibration for image processing. Moreover, they employ the “best scene on top” (BOT) technique in the mosaicking process of PlanetScope imagery~\citep{CoralAtlas-paper}.  {In Step b, we check what type of reef community mapping is desired by the user, i.e., benthic or geomorphic mapping.}   We then create a geomorphic map of the region, where the bathymetric data (Step c) are concatenated with the imagery obtained in the previous step.
Next, we evaluate  the clustering algorithms (Figure~\ref{fig:flowchart}---Step d)   to create clustering regions (segments). We need to evaluate the results and thus need a way to ensure that the acquired segments are meaningful. Hence, we apply   qualitative analysis, where we assign each cluster  a colour according to the map  used for comparison; then, we compare  the maps qualitatively  (visually), side by side (Figure~\ref{fig:flowchart}---Step e). This helps in assigning the labels to the clusters based on the visual similarities to the existing maps. If the results obtained are unsatisfactory, the clustering algorithm is again applied to the data with new parameters (Step g).  The final step incorporates map refinement and clean-up (Step h), wherein we merge the extra clusters with the closest region of interest ~\citep{CoralAtlas-paper} to generate the coral reef map of interest (Step i).

%  apply logical rules incorporating the geomorphology of the region   and remap the smaller clusters  along with the bigger clusters based on field knowledge and logical rules

 Map refinement by logical rules (Step h) remaps the smaller, excess clusters to the major cluster of a given label surrounding them. This ensures that the smaller clusters are merged to obtain a refined map with only the regions having labels of interest.

\section{Results}\label{sec3} 

\subsection{K-Means and GMM Clustering}

\begin{figure}[htbp!]
 
 \begin{tabular}{c}
 \includegraphics[width=7cm] {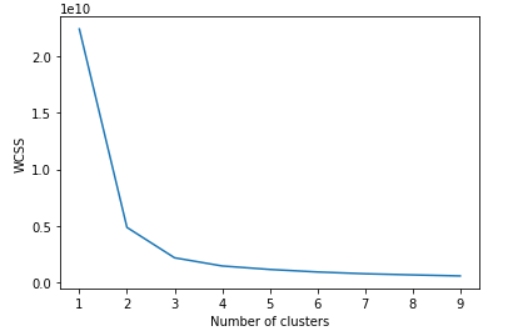} \\
 (\textbf{a}) WCSS\\
 \includegraphics[width=7cm]{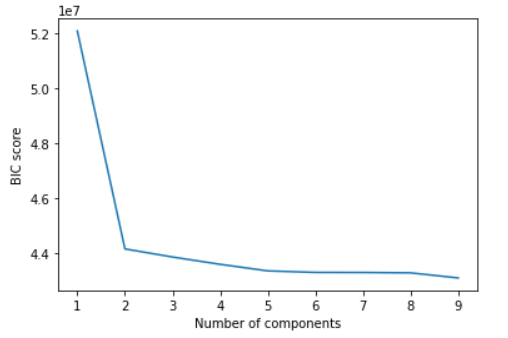} \\
 (\textbf{b}) BIC\\
\end{tabular}
%   %  \centering
%     \begin{subfigure}{0.6\textwidth}
%       % \centering
%         \includegraphics[width=7cm] {Elbow_Kmeans_Benthic.PNG}
%      %   \captionsetup{justification=centering}
%             \caption{WCSS}
%                  \end{subfigure}
%     \vspace{12pt}
%     
%     \begin{subfigure}{0.6\textwidth}
%    %     \centering
%         \includegraphics[width=7cm]{BIC_GMM_Benthic.PNG}
%     {    \caption{BIC} }
%     \end{subfigure} 
        \caption{ {The} %MDPI: Please change the terms into scientific notations in the figure,  e.g., "$8 \times 10^{3}$", not "8E3". This is how the python library produces  the fig and we cannot change it. We prefer to leave as it is. 
 elbow method to evaluate the number of clusters (components) for k-means using WCSS and BIC, respectively.}
        \label{fig:Evaluation_methods}
\end{figure}

We begin by finding the optimal number of clusters for {our framework's respective clustering methods, i.e., k-means and GMM.}   In the case of k-means clustering, we use the elbow method, which  plots the sum of the square distance to find the number of optimal clusters ($k$ value) by calculating the distance between a data point and the cluster (WCSS). The point where the curve starts to flatten and resembles the elbow of the curve is chosen as the $k$ value. {In Figure~\ref{fig:Evaluation_methods}, an elbow can be seen at $k=3$. In the case of GMM, we use the {Bayesian information criterion} (BIC) to find the value of $k$. The gradient of the BIC score curve, much similar to finding the elbow of the curve, is used to estimate  the optimal number of clusters for the data. A lower score indicates that the model better fits the data. However, in order to avoid over-fitting, this technique penalises the methods with a large number of clusters. In Figure~\ref{fig:Evaluation_methods}, we select the point that reflects the major change in the gradient which can be seen at $k=2$.}

\begin{figure}[htbp!]
 
 \begin{adjustwidth}{-\extralength}{1.3cm}
     \centering
     \begin{subfigure}[b]{0.5\textwidth}
         \centering
         \includegraphics[width=8.3cm]{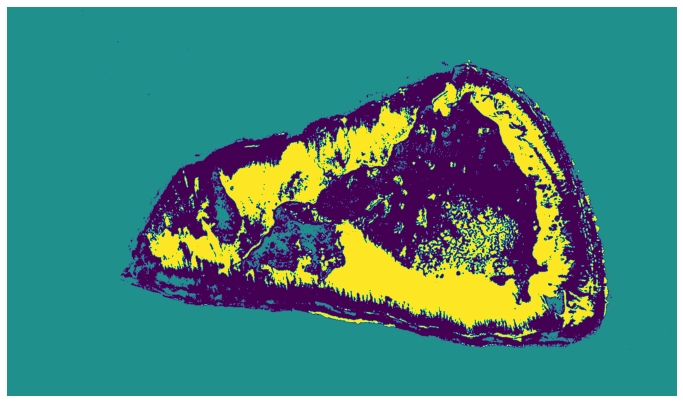}\\
     {\captionsetup{justification=centering}    \caption{k = 3} }
     \end{subfigure}
     \hfill
     \begin{subfigure}[b]{0.5\textwidth}
         \centering
         \includegraphics[width=8.2cm]{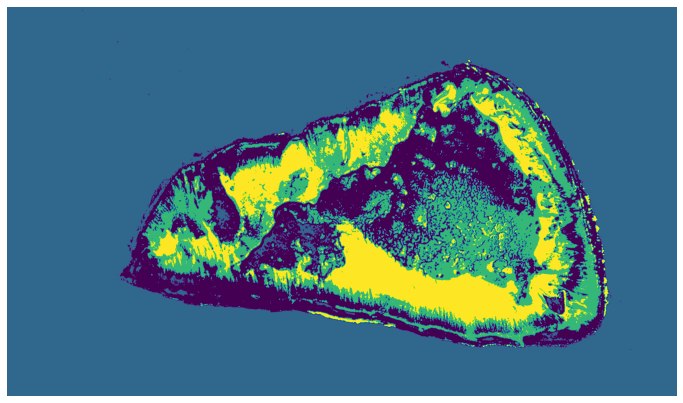}\\
   {\captionsetup{justification=centering}      \caption{k = 4} }\vspace{4pt}
     \end{subfigure} 
     
\hfill
     \begin{subfigure}[b]{0.5\textwidth}
         \centering
         \includegraphics[width=8.2cm]{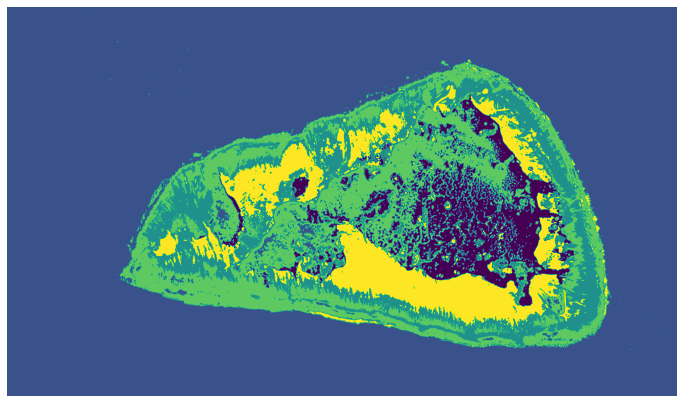}
      {\captionsetup{justification=centering}   \caption{k = 5} }
     \end{subfigure} 
     \end{adjustwidth}
        \caption{Evaluating $k$ in k-means clustering.}
        \label{fig:Eval_k}
\end{figure}

In our case, we provide a visual comparison to evaluate the methods, i.e. qualitative comparison of the clustered maps of the reef and the Allen Coral Atlas. Hence, while keeping the   elbow method and BIC score in mind, we review the results obtained using the clustering methods and choose $k$ based on maximum resemblance to the  regions of interest in the coral maps (e.g., Figure~\ref{fig:Eval_k}b).

%evaluate k

\subsection{Comparison of Selected Clustering Results}

\begin{figure}[htbp!]

\begin{adjustwidth}{-\extralength}{1.2cm}
\centering %% If there is a figure in wide page, please release command \centering

     \centering
     \begin{subfigure}[b]{0.5\textwidth}
         \centering
         \includegraphics[width=8.1cm]{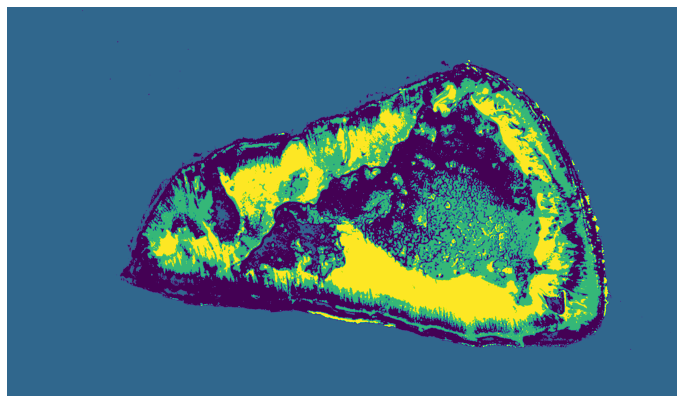}
        {\captionsetup{justification=centering} \caption{K-means} }
     \end{subfigure}
     \hfill
     \begin{subfigure}[b]{0.5\textwidth}
         \centering
         \includegraphics[width=8.1cm]{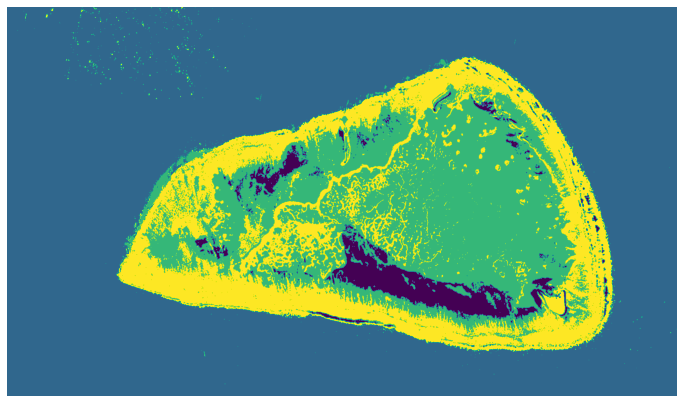}
        {\captionsetup{justification=centering} \caption{GMM} }
     \end{subfigure} 
          \hfill
     \begin{subfigure}[b]{0.5\textwidth}
         \centering
         \includegraphics[width=8.1cm]{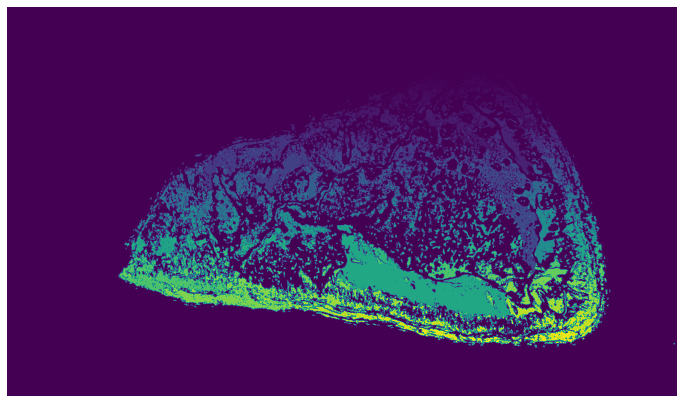}
      {\captionsetup{justification=centering}   \caption{DBSCAN} }
     \end{subfigure} 
          \hfill
     \begin{subfigure}[b]{0.5\textwidth}
         \centering
         \includegraphics[width=8.1cm]{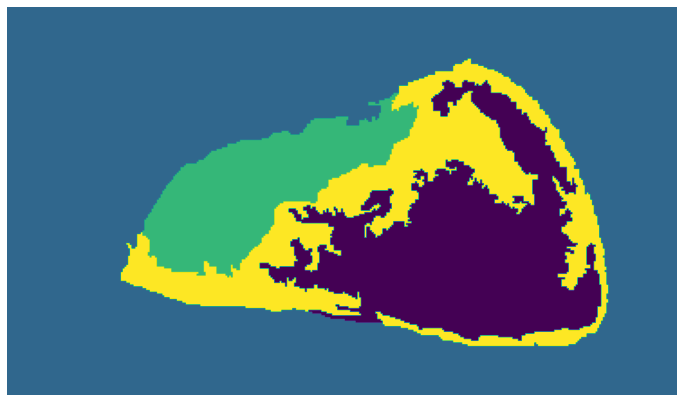}
   {\captionsetup{justification=centering}      \caption{HAC} }
     \end{subfigure} 
  \end{adjustwidth}
        \caption{ {Benthic} %MDPI: for figures 6-8, we changed the images to two row two column, please confirm the format.  OK
 maps produced with the four  different clustering methods. }
        \label{fig:Clustering_results_4methods_Benthic}
\end{figure}

\begin{figure}[htbp!]
  
 \begin{adjustwidth}{-\extralength}{1.2cm}
     \centering
     \begin{subfigure}[b]{0.5\textwidth}
         \centering
         \includegraphics[width=8.2cm]{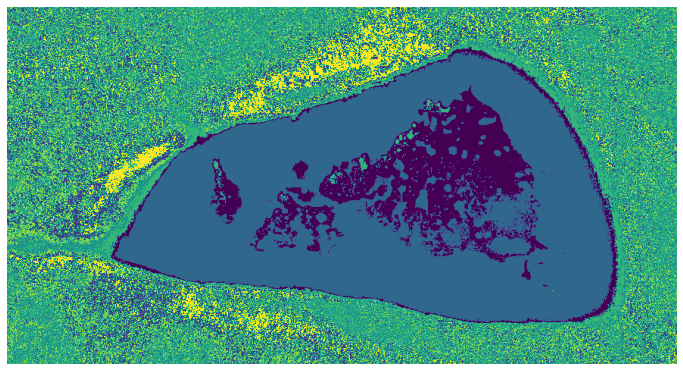}
     {\captionsetup{justification=centering}    \caption{K-means} }
     \end{subfigure}
     \hfill
     \begin{subfigure}[b]{0.5\textwidth}
         \centering
         \includegraphics[width=8.2cm]{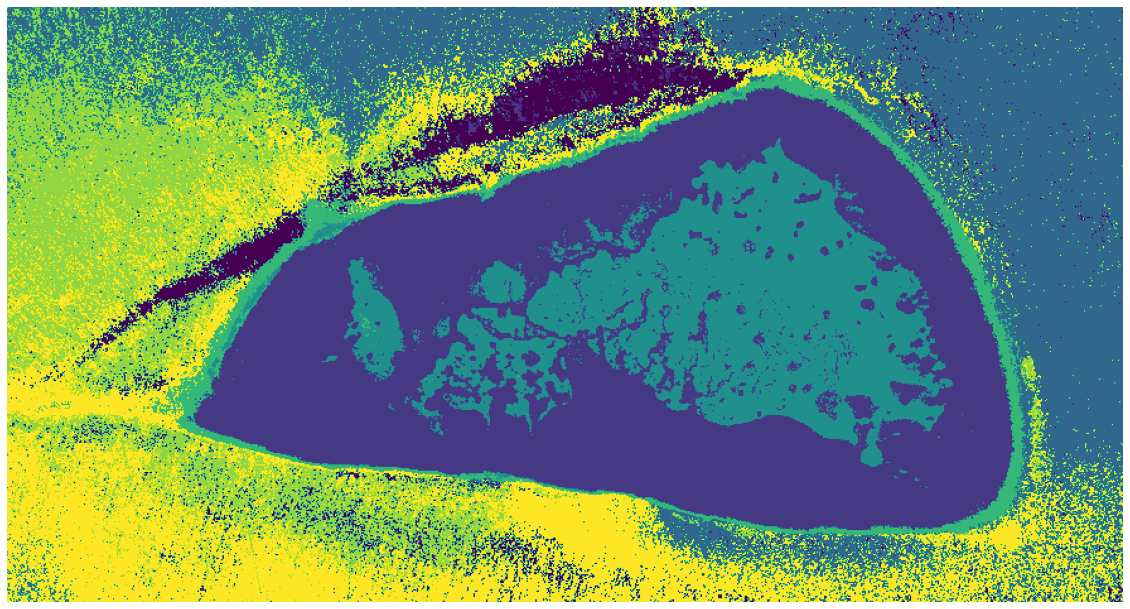}
      {\captionsetup{justification=centering}   \caption{GMM} }
     \end{subfigure} 
          \hfill
     \begin{subfigure}[b]{0.5\textwidth}
         \centering
         \includegraphics[width=8.2cm]{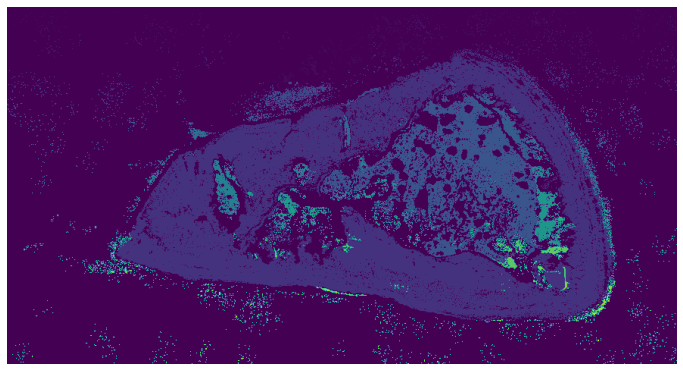}
    {\captionsetup{justification=centering}     \caption{DBSCAN} }
     \end{subfigure} 
          \hfill
     \begin{subfigure}[b]{0.5\textwidth}
         \centering
         \includegraphics[width=8.2cm]{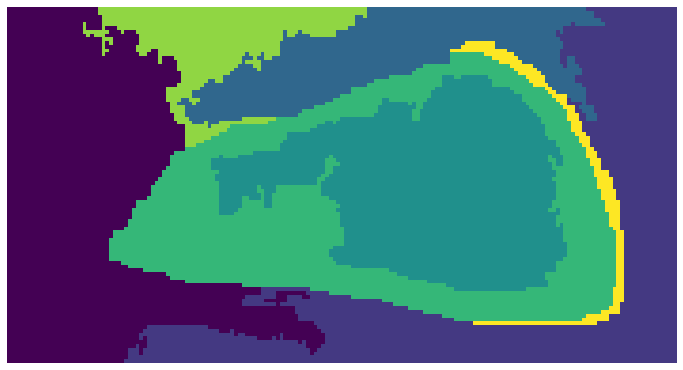}
       {\captionsetup{justification=centering}  \caption{HAC} }
     \end{subfigure} 
     \end{adjustwidth}
        \caption{Geomorphic maps produced with the four  different clustering methods. }
        \label{fig:Clustering_results_4methods_Geomorphic}
\end{figure}

Figures \ref{fig:Clustering_results_4methods_Benthic} and \ref{fig:Clustering_results_4methods_Geomorphic} represent the benthic and geomorphic maps generated with the four clustering methods considered in our framework. Agglomerative clustering is computationally exhaustive; hence, we  down-scaled the data to 20\% of the original size, which led to the loss of finer details of the areas of interest (Figures~\ref{fig:Clustering_results_4methods_Benthic}d and \ref{fig:Clustering_results_4methods_Geomorphic}d).  DBSCAN gave an adequate result, especially for the geomorphic map (Figure~\ref{fig:Clustering_results_4methods_Benthic}c) of the region, by removing the bathymetric noise and focusing on the reef area. However, a large number of  small clusters are not ideal for the map refinement step ahead for visual comparison with existing maps. 
The k-means results are given in Figures ~\ref{fig:Clustering_results_4methods_Benthic}a and \ref{fig:Clustering_results_4methods_Geomorphic}a. The GMM results given in \mbox{Figures  \ref{fig:Clustering_results_4methods_Benthic}b and \ref{fig:Clustering_results_4methods_Geomorphic}b} are satisfactory results in both benthic and geomorphic mapping compared  with DBSCAN and AGNES. 

\subsection{Comparison with Allen Coral Atlas}

\begin{figure}[htbp!]
  
 \begin{adjustwidth}{-\extralength}{1.4cm}
     \centering
     \begin{subfigure}{0.5\textwidth}
         \centering
         \includegraphics[width=8.2cm]{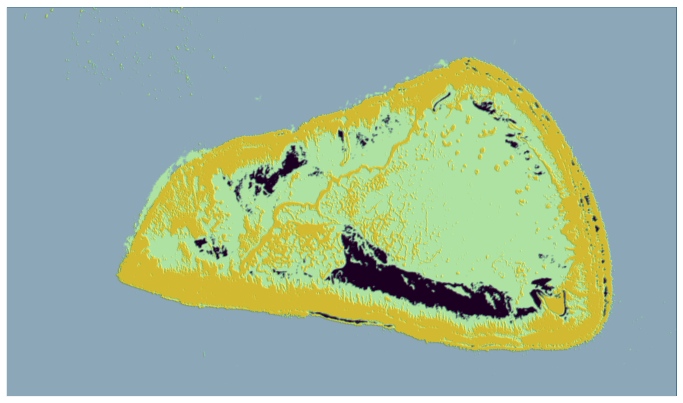}
       {\captionsetup{justification=centering}  \caption{} }
     \end{subfigure}
     \hfill
     \begin{subfigure}{0.5\textwidth}
         \centering
         \includegraphics[width=8.2cm]{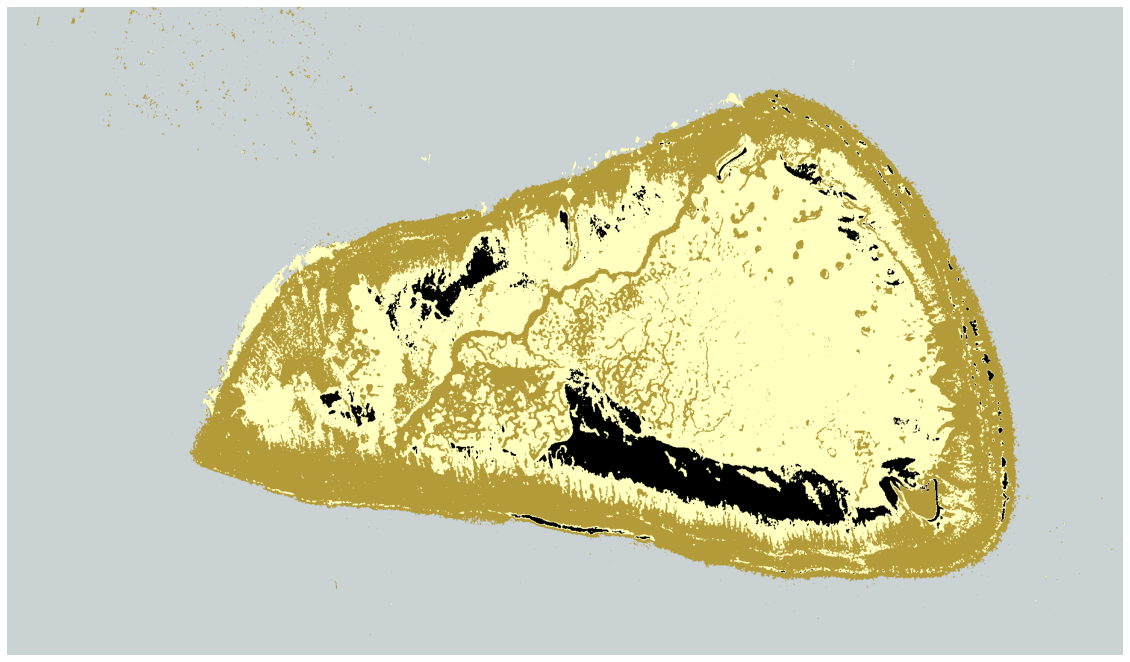}
       {\captionsetup{justification=centering}  \caption{}} \vspace{5pt}
     \end{subfigure} 
          \hfill
     \begin{subfigure}{0.5\textwidth}
         \centering
         \includegraphics[width=8.2cm]{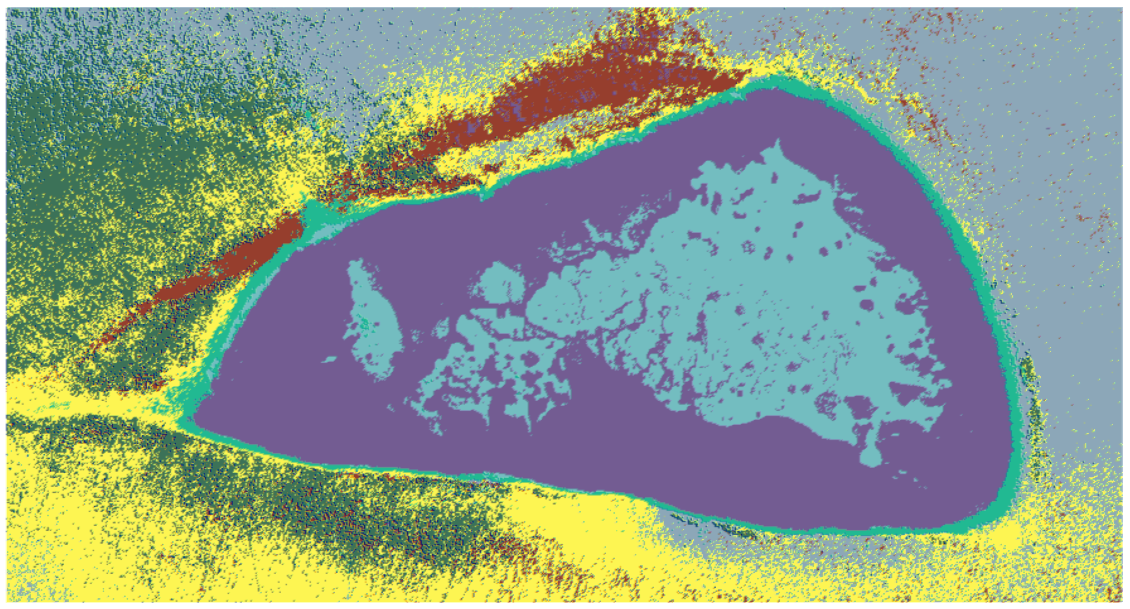}
     {\captionsetup{position=bottom,justification=centering}    \caption{} }\vspace{5pt}
     \end{subfigure} 
          \hfill
     \begin{subfigure}{0.5\textwidth}
         \centering
         \includegraphics[width=8.2cm]{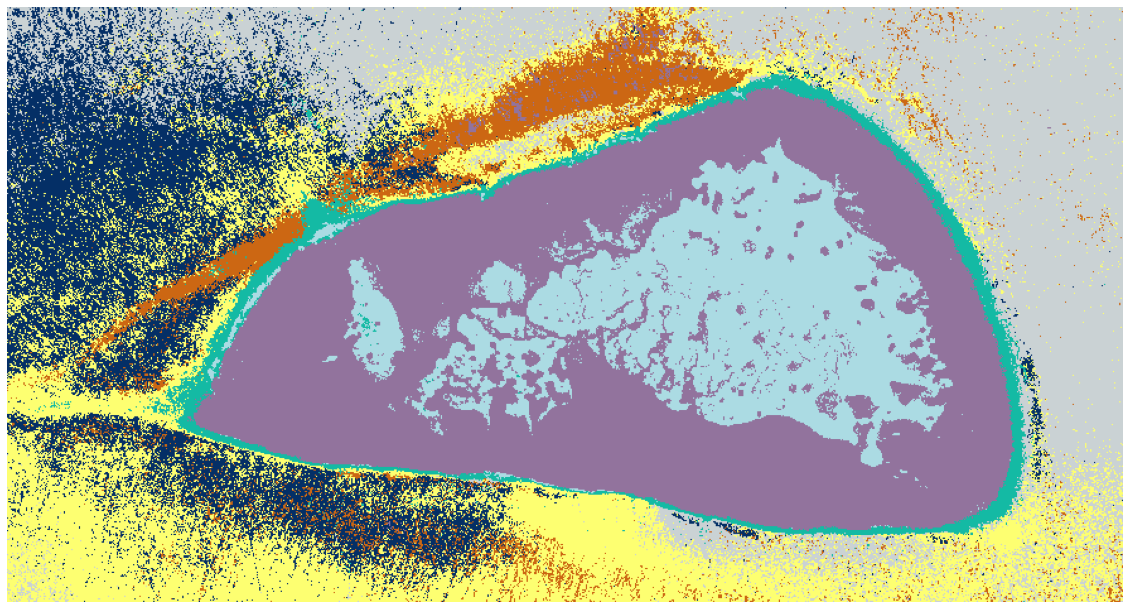}
      {\captionsetup{justification=centering}      \caption{} }
     \end{subfigure} 
     \end{adjustwidth}
    \caption{ {Preliminary} %MDPI: for figures 8-10, we moved the subfigure explantion to figure caption, please confirm it. OK
 coral map results---GMM. (\textbf{a}) Benthic map preliminary result overlay; \mbox{(\textbf{b}) benthic} map preliminary result; (\textbf{c}) geomorphic map preliminary result overlay; (\textbf{d}) geomorphic map preliminary result overlay.}
    \label{fig:aca_preliminary_res}
\end{figure}

\begin{figure}[htbp!]

 \begin{adjustwidth}{-\extralength}{1.2cm}
     \centering
     \begin{subfigure}[b]{0.5\textwidth}
         \centering
         \includegraphics[width=8.2cm]{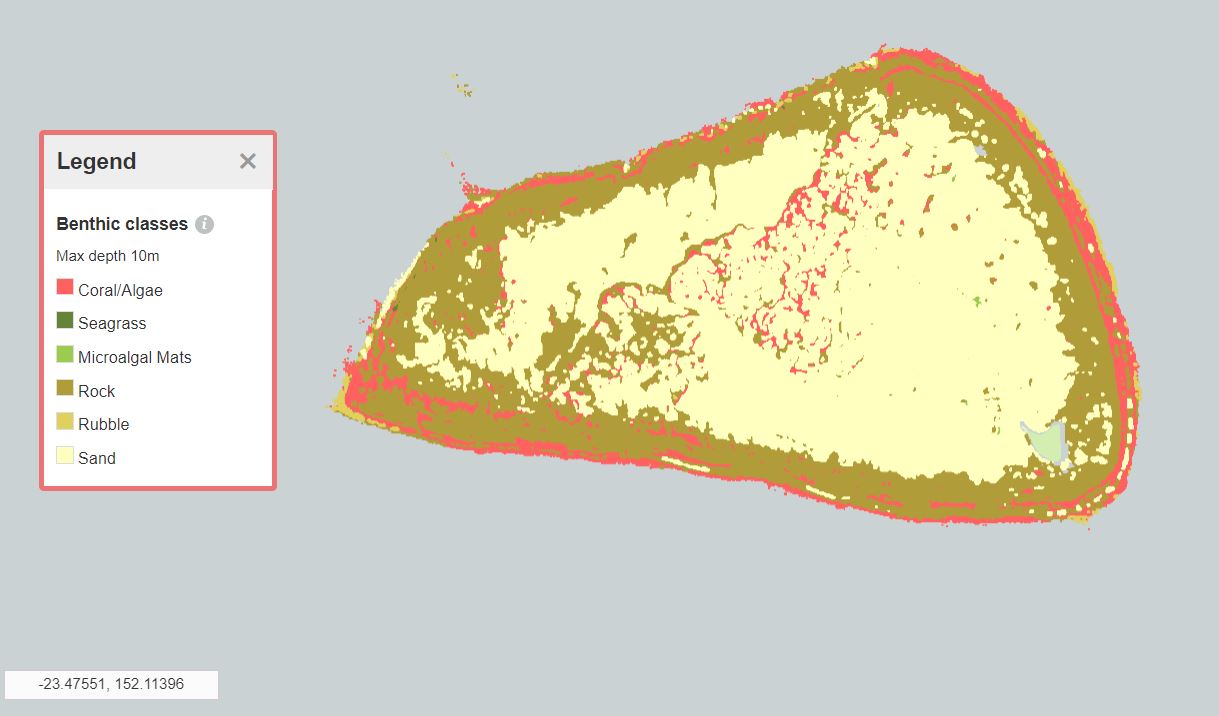}
         {\captionsetup{justification=centering}\caption{} }
     \end{subfigure} 
     \hfill
     \begin{subfigure}[b]{0.5\textwidth}
         \centering
         \includegraphics[width=8.2cm]{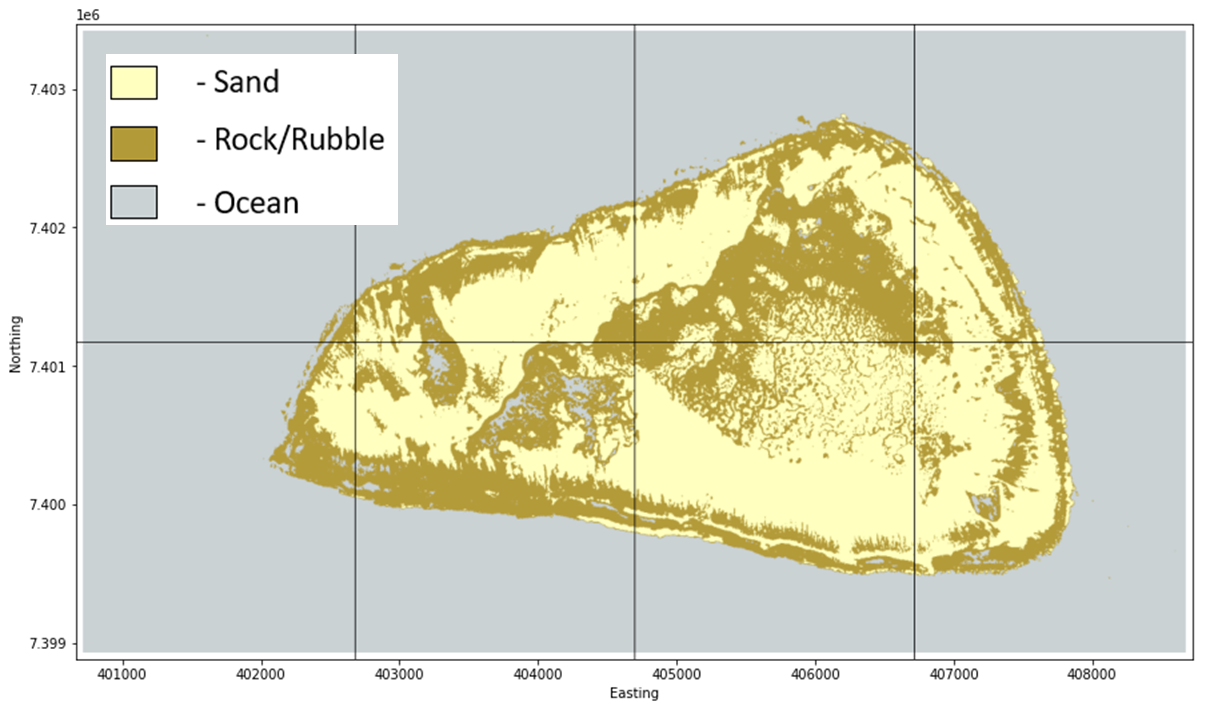}
       {\captionsetup{justification=centering}  \caption{} }\vspace{5pt}
     \end{subfigure} 
     
     \hfill
     \begin{subfigure}[b]{0.5\textwidth}
         \centering
         \includegraphics[width=8.2cm]{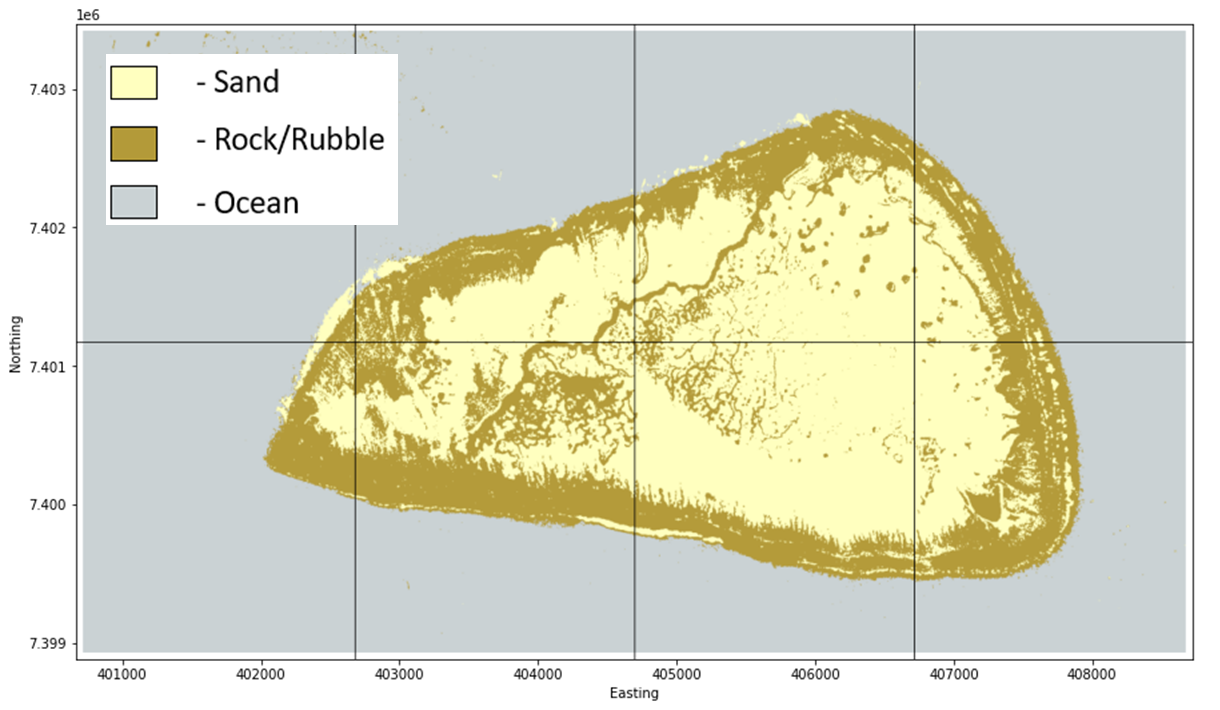}
  {\captionsetup{justification=centering}       \caption{} }
     \end{subfigure} 
\end{adjustwidth}
\caption{ {Benthic} %MDPI: 1. Please change the terms into scientific notations in the figure,  e.g., "$8 \times 10^{3}$", not "8E3". 2. Please use commas to separate thousands for numbers with five or more digits (not four digits) in the picture, e.g., "10000" should be "10,000".   We prefer to keep it as it is since this is how the python library produced the figure. 
 map results---Allen Coral Atlas. (\textbf{a}) Benthic map from the Allen Coral Atlas~\citep{CoralAtlas-paper};  \mbox{(\textbf{b})  k-means} benthic map; (\textbf{c}) GMM benthic map.}
\label{fig:Benthic_ACA}
\end{figure}

Next, we execute k-means and GMM clustering  methods  with a selected number of clusters ($k=4$) that gives maximum resemblance to actual data. In the GMM, we set the   covariance type parameter set to full to ensure  that each component of the GMM had its own general covariance matrix. The preliminary results obtained (Figure~\ref{fig:aca_preliminary_res}a) represent the four clusters. We observe that the GMM clustering method has well extracted the flat sand region represented by the black cluster. We then combine the clustered region and the sand region to obtain the final result. The clustering results of  generating benthic coral maps obtained with k-means  (Figure~\ref{fig:Benthic_ACA}b) and the GMM  (Figure~\ref{fig:Benthic_ACA}c) showcase three clusters, namely, ocean, sand, and rock/rubble. Upon visual comparison with the benthic map from the Allen Coral Atlas (Figure~\ref{fig:Benthic_ACA}a), we can see  that GMM  provides clusters with higher similarity than  k-means clustering.  {Figure}~\ref{fig:aca_preliminary_res} shows the results of the logical rules (Figure~\ref{fig:flowchart}---Step h) used to create  the benthic map generated using GMM. Figure~\ref{fig:aca_preliminary_res}b has a black cluster that got remapped to a sand (yellow) cluster in the map refinement stage ( {Figure}~\ref{fig:Benthic_ACA}c). %MDPI: wrong figure order, figures 6,7 should be cited after figure 5, please revise it to the first citation of all figures in numberical order.   THIS paragraph was moved from earlier section, now figures are cited as needed

In the case of the geomorphic map, we set the number of clusters ($k-7$) for both methods, i.e. k-means and GMM. The preliminary results obtained using the GMM (Figure~\ref{fig:aca_preliminary_res}b)  provide extra clusters in the ocean region by making a distinction in water bathymetry. We combine the clusters in the ocean region by visually comparing them with the Allen Coral Atlas geomorphic map.

\begin{figure}[H]

 \begin{adjustwidth}{-\extralength}{1.2cm}
     \centering
     \begin{subfigure}[b]{0.5\textwidth}
         \centering
         \includegraphics[width=8.2cm]{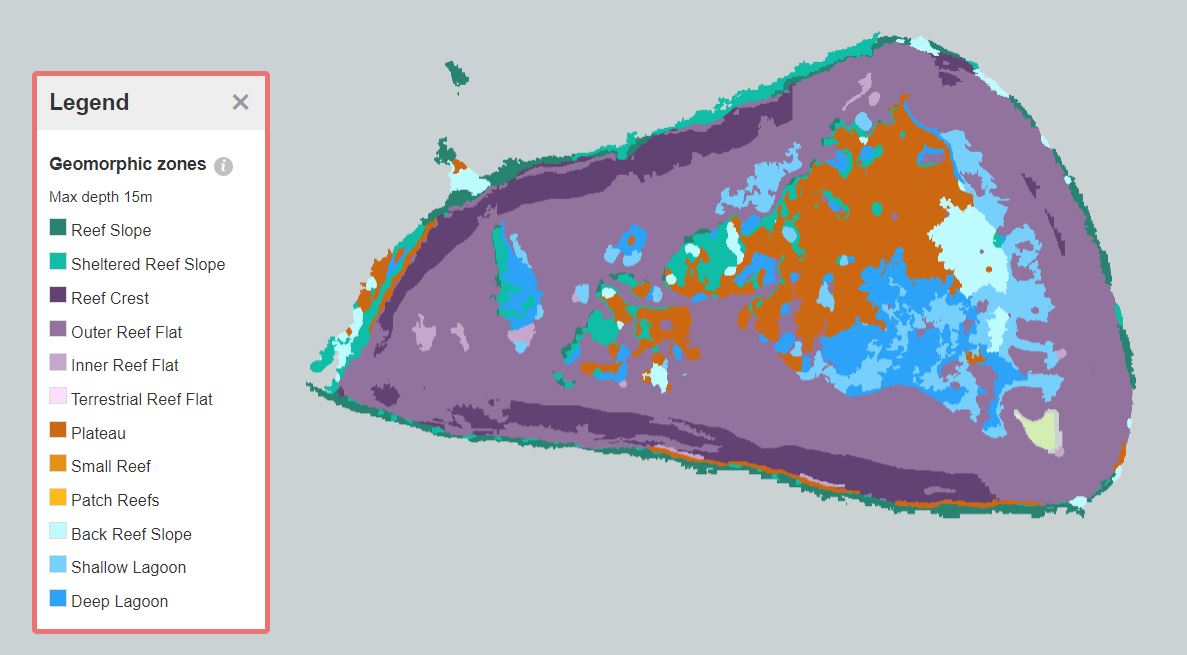}
       \captionsetup{justification=centering}  \caption{} 
     \end{subfigure} 
     \hfill
     \begin{subfigure}[b]{0.5\textwidth}
         \centering
         \includegraphics[width=8.2cm]{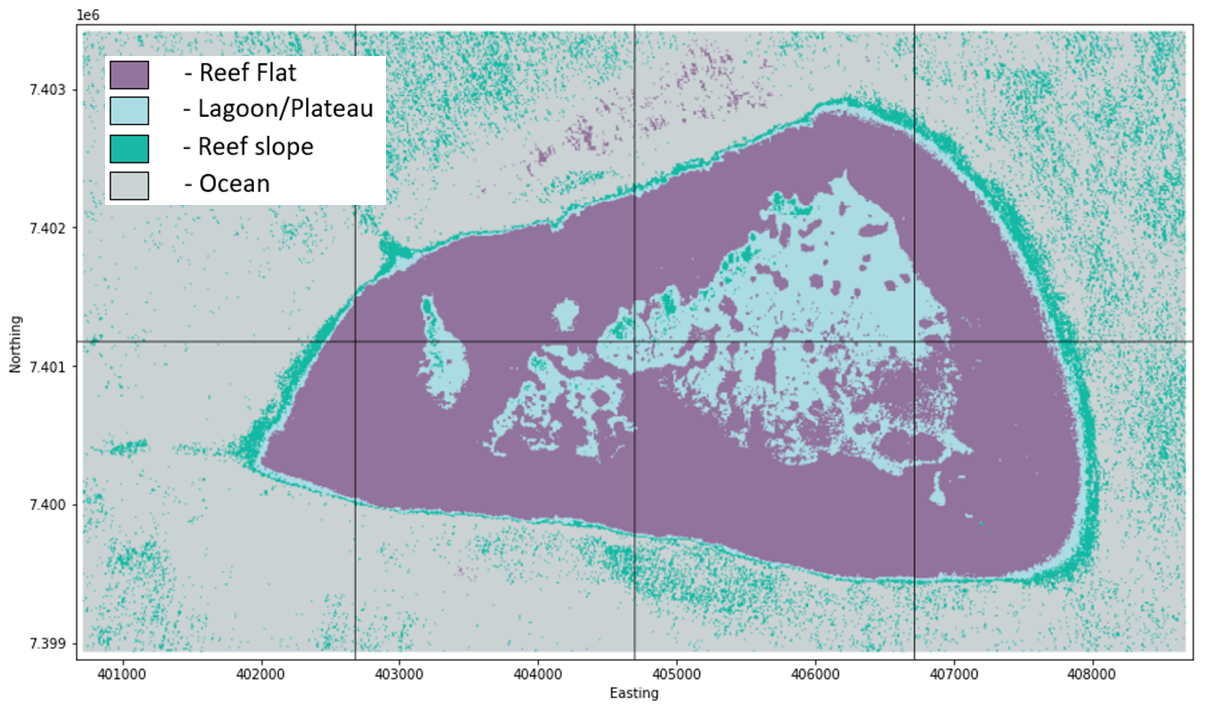}
      {\captionsetup{justification=centering}   \caption{} }\vspace{5pt}
     \end{subfigure} 
     
     \hfill
     \begin{subfigure}[b]{0.5\textwidth}
         \centering
         \includegraphics[width=8.2cm]{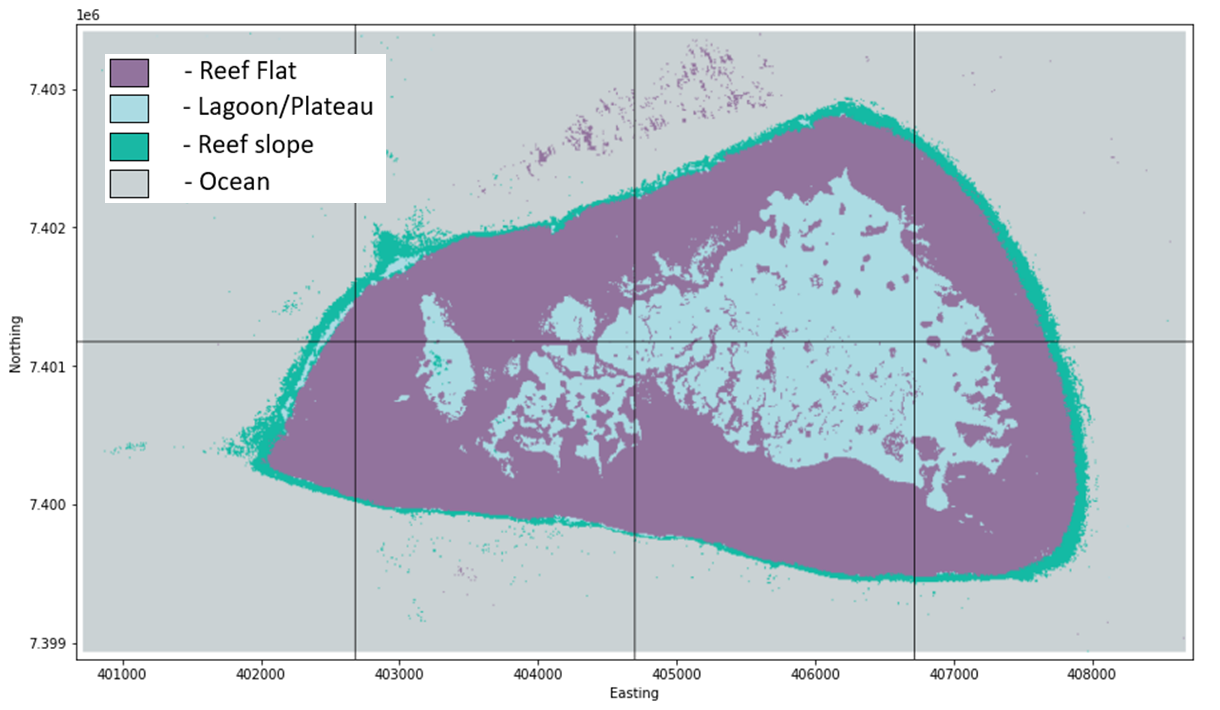}
   {\captionsetup{justification=centering}      \caption{} }
     \end{subfigure} 
     \end{adjustwidth}
\caption{ {Geomorphic} %MDPI: 1. Please change the terms into scientific notations in the figure,  e.g., "$8 \times 10^{3}$", not "8E3". 2. Please use commas to separate thousands for numbers with five or more digits (not four digits) in the picture, e.g., "10000" should be "10,000". We prefer to keep it as it is. 
 map results---Allen Coral Atlas. (\textbf{a})  Geomorphic map from Allen Coral Atlas~\citep{CoralAtlas-paper}; (\textbf{b}) k-means geomorphic map;  (\textbf{c}) GMM geomorphic map.}
\label{fig:Geo_ACA}
\end{figure}

\textls[-15]{We observe that the final geomorphic maps generated using k-means  (Figure~\ref{fig:Geo_ACA}b) and  GMM  (\mbox{Figure~\ref{fig:Geo_ACA}c})} generate four clusters: reef flat, lagoon/ plateau, reef slope, and ocean. The reef flat and the lagoon/ plateau region created with the GMM had a greater resemblance to the original geomorphic map given by the Allen Coral Atlas  (Figure~\ref{fig:Geo_ACA}a). A general limitation in clustering methods  for reef habitat mapping   is  the classification of  local regions which is due to a lack of labelled data. Nevertheless, this approach can be useful for gathering a basic overview of reef habitats  without the need for manual labelling, which is a labour-intensive task.

\vspace{-12pt}

  \vspace{-6pt}
    
 \section{Discussion} \label{sec4}
 In this study, we presented a framework that processes remote sensing data and compared four clustering methods (k-means, GMM, AGNES, and DBSCAN) for generating reef habitat maps using remote sensing data of the One Tree Island reef.  We utilised the Allen Coral Atlas mosaic of the region to generate benthic maps using clustering methods and incorporated bathymetric data for understanding the geomorphic habitats of the region.   We selected the appropriate clustering method based on visual comparative analysis to ensure that the clustered regions took into account the local field knowledge and geology. Thus, we generated a three-class benthic map distinguishing among sand, rock/rubble, and ocean, and a four-class geomorphic map identifying reef flat, reef slope, lagoon, and ocean regions on the map.    The quality of segmentation depended on parameter tuning and  the choice of the clustering method. 
 
The coral maps obtained with the proposed framework can be considered  preliminary maps for understanding the geomorphology of the region of interest. Our   framework can provide additional support to the supervised methods that are mainly used for reef mapping, such as the one used in the Allen Coral Atlas. In future studies,  our  framework can  be used to understand more about surface geology in coral reefs, given the sea-level rise for thousands of years. Since there are drilled reef-core data of the site that give insights into the evolution of coral reef structures for thousands of years, this can help extend software such as BayesReef~\citep{pall2020bayesreef}, which uses Bayesian inferences for stimulating one-dimensional reef cores into three-dimensional reef evolution simulations.  
 
In terms of the limitations of the framework, we  note that there is a  lengthy process of taking into account the visual comparison to find the optimal parameters for reef mapping. Moreover, it is also difficult to assess the accuracy of clusters on their own, and field knowledge and labelled maps to allocate labels to each clustering region are required. 
However, the study has revealed certain combinations of hyper-parameters ($k$ value) that are useful for reef areas, and the same can be used in future studies in which the framework is applied to other regions. Furthermore, the current study considered a relatively small area, and it can be a challenge if clustering methods are used on a large area, such as the entire GBR region. The framework would then need to be extended using a distributed/parallel computing infrastructure so that the method can work with smaller regions, i.e., large regions can be divided using a grid and the results can be combined.

We note that we did not consider data with  temporal dimension, as they were not available in our dataset. In the future, given the availability of temporal satellite data, our framework could be extended to evaluate decadal changes in reef habitat mapping that can capture extreme climate events,  such as cyclones, and other events, such as tsunamis, which have devastating impacts on coral reef systems \cite{sublime2019automatic}. In such a study, the need for parameter tuning and the evaluation of clustering methods can be eliminated using the results from this study. Temporal data can also be used to study short-term seasonal changes in reef habitats. The quality of segmentation depends on parameter tuning and the choice of the clustering method. Our paper evaluates different clustering algorithms and recommends the best for this problem. The method is replicable and readability available with the results of the study and the availability of code and data.

\section{Conclusions and Future Work}\label{sec5}
 
In this study, we presented a framework to compare  different clustering methods for the task of reef habitat mapping using unlabelled remote sensing data. We used   One Tree Island of the GBR to demonstrate the effectiveness of the framework. The framework transformed the raw clusters into a reef habitat map using field knowledge and map refinement operations based on logical rules that were gathered from expert knowledge. The results show that the k-means and GMM clustering methods are the most suitable  for benthic and geomorphic reef mapping, as these methods created the maps that were the most visually similar to the maps obtained using related methods (Coral Atlas).

In future work, our framework can be used for reef change detection, especially when field inspection cannot be easily conducted; e.g., in case of natural disasters such as  tsunamis, storms, and  cyclones. The framework can help assess the impact of extreme climate events (cyclones and storms) on reef habitats, which  can play a crucial role in reef restoration projects.
Furthermore, the framework can also be utilised for generating maps using remote sensing data of the regions for which labelled data are unavailable, such as remote sensing data obtained from  Mars and Moon exploration projects. Our framework is a way to address the challenges faced by reef scientists that involve finding labelled data for analysis and the need for manually labelling reef regions, especially in large regions. It can be considered a low-cost and robust approach to working with raw data during the exploration stage of a research study. In future work, our framework can be extended with other clustering methods and further validated using regions with labelled reef data.

\vspace{6pt}
%%%%%%%%%%%%%%%%%%%%%%%%%%%%%%%%%%%%%%%%%%
\authorcontributions{S. Barve contributed to programming,  experiments, and writing. J. M. Webster provided data and contributed to analyses and writing.   R. Chandra conceptualised and supervised the project and contributed to writing and analyses. All authors have read and agreed to the published version of the manuscript.}

\funding{This research received no external funding}
 
\dataavailability{ We provide open source code and data for our proposed framework at GitHub  {repository}  (accessed on 1 June 2023).
    \url{https://github.com/DARE-ML/reefhabitat-mapping}.}

\conflictsofinterest{The authors declare no conflict of interest.} 

\begin{adjustwidth}{-\extralength}{0cm}
%\centering %% If there is a figure in wide page, please release command \centering

 \reftitle{References}

%\bibliographystyle{sn-basic} 

%\bibliographystyle{sn-basic}

%\bibliographystyle{elsarticle-harv}
%\bibliography{2017,2018,Chandra-Rohitash,Bays,sample,cnnbayes,remotesensing} 
\PublishersNote{}
\end{adjustwidth}
\end{document}